\documentclass[letterpaper, 10 pt, conference]{ieeeconf}  

\IEEEoverridecommandlockouts  

\usepackage{times}

\usepackage{cite}
\usepackage{multicol}
\usepackage{graphics} 
\usepackage{amsmath} 
\usepackage{amssymb}  
\usepackage[dvipsnames]{xcolor}
\usepackage{hyperref}
\hypersetup{
    colorlinks=true,
    linkcolor=black,    
    urlcolor=black,
    citecolor=black
    }

\usepackage[ruled,vlined,linesnumbered]{algorithm2e}
\usepackage{epsfig} 
\usepackage{times} 
\usepackage{amsmath} 
\usepackage{amssymb}  
\usepackage{multirow}
\usepackage{algpseudocode}
\usepackage{subfigure}

\usepackage{microtype}

\newcommand{\GGG}{\mathcal{G}}

\newcommand{\EEE}{\mathcal{E}}

\newcommand{\VVV}{\mathcal{V}}

\newcommand{\OOO}{\mathcal{O}}

\newcommand{\NNN}{\mathcal{N}}

\newcommand{\MMM}{\mathcal{M}}

\newcommand{\SIM}{\mathbf{S}}
\newcommand{\sss}{\mathbf{s}}

\newcommand{\hh}{\mathbf{h}}

\newcommand{\vv}{\mathbf{v}}
\newcommand{\qq}{\mathbf{q}}

\newcommand{\AAAA}{\mathbf{A}}
\newcommand{\DD}{\mathbf{D}}

\newcommand{\mm}{\mathbf{m}}
\newcommand{\bb}{\mathbf{b}}

\newcommand{\WW}{\mathbf{W}}

\newcommand{\zz}{\mathbf{z}}
\newcommand{\YY}{\mathbf{Y}}

\newcommand{\RR}{\mathcal{R}}

\title{\Large \bf Bandwidth-Adaptive Spatiotemporal Correspondence Identification \\ for Collaborative Perception}

\usepackage{authblk}

\author{
\thanks{*This work was partially supported by  NSF CAREER Award IIS-2308492, DARPA Young Faculty Award (YFA) D21AP10114-00, and DEVCOM ARL A2I2 CRA W911NF-23-2-0005.}
  Peng Gao\textsuperscript{1}\thanks{\textsuperscript{1}North Carolina State University, email: \texttt{pgao5@ncsu.edu}},
  Williard Joshua Jose\textsuperscript{2}\thanks{\textsuperscript{2}Human-Centered Robotics Lab at University of Massachusetts Amherst, email: \texttt{$\{$wjose, hao.zhang$\}@$umass.edu}},
  Hao Zhang\textsuperscript{2}
}

\begin{document}

\maketitle

\begin{abstract}
Correspondence identification (CoID) is an essential capability in multi-robot collaborative perception,
which enables a group of robots to consistently refer to the same objects within their respective fields of view.
In real-world applications, such as connected autonomous driving, vehicles face challenges in directly sharing raw observations due to limited communication bandwidth.
In order to address this challenge,  we propose a novel approach for bandwidth-adaptive spatiotemporal CoID in collaborative perception.
This approach allows robots to progressively select partial spatiotemporal observations and share with others, 
while adapting to communication constraints that dynamically change over time.
We evaluate our approach across various scenarios in connected autonomous driving simulations.
Experimental results validate that our approach enables CoID and adapts to dynamic communication bandwidth changes.
In addition, our approach achieves $8\%$-$56\%$ overall improvements in terms of covisible object retrieval for CoID and data sharing efficiency, which outperforms previous techniques and achieves the state-of-the-art performance. More information is available at: \url{https://gaopeng5.github.io/acoid}.
\end{abstract}

\section{Introduction}

Multi-robot systems have earned significant attention over the past few decades, primarily owing to their reliability and efficiency in tackling cooperative tasks. These tasks encompass a broad spectrum, including collaborative manufacturing \cite{gao2021bayesian, zhang2020real}, multi-robot search and rescue \cite{reily2021adaptation, reily2021balancing}, and connected autonomous driving \cite{guo2019collaborative}.
To facilitate effective collaboration among robots,
a fundamental capability is collaborative perception. This capability allows multiple robots to share perceptual data about their surrounding environment, thereby fostering a shared situational awareness.

\begin{figure}[t]
\vspace{6pt}
\centering
\includegraphics[width=0.458\textwidth]{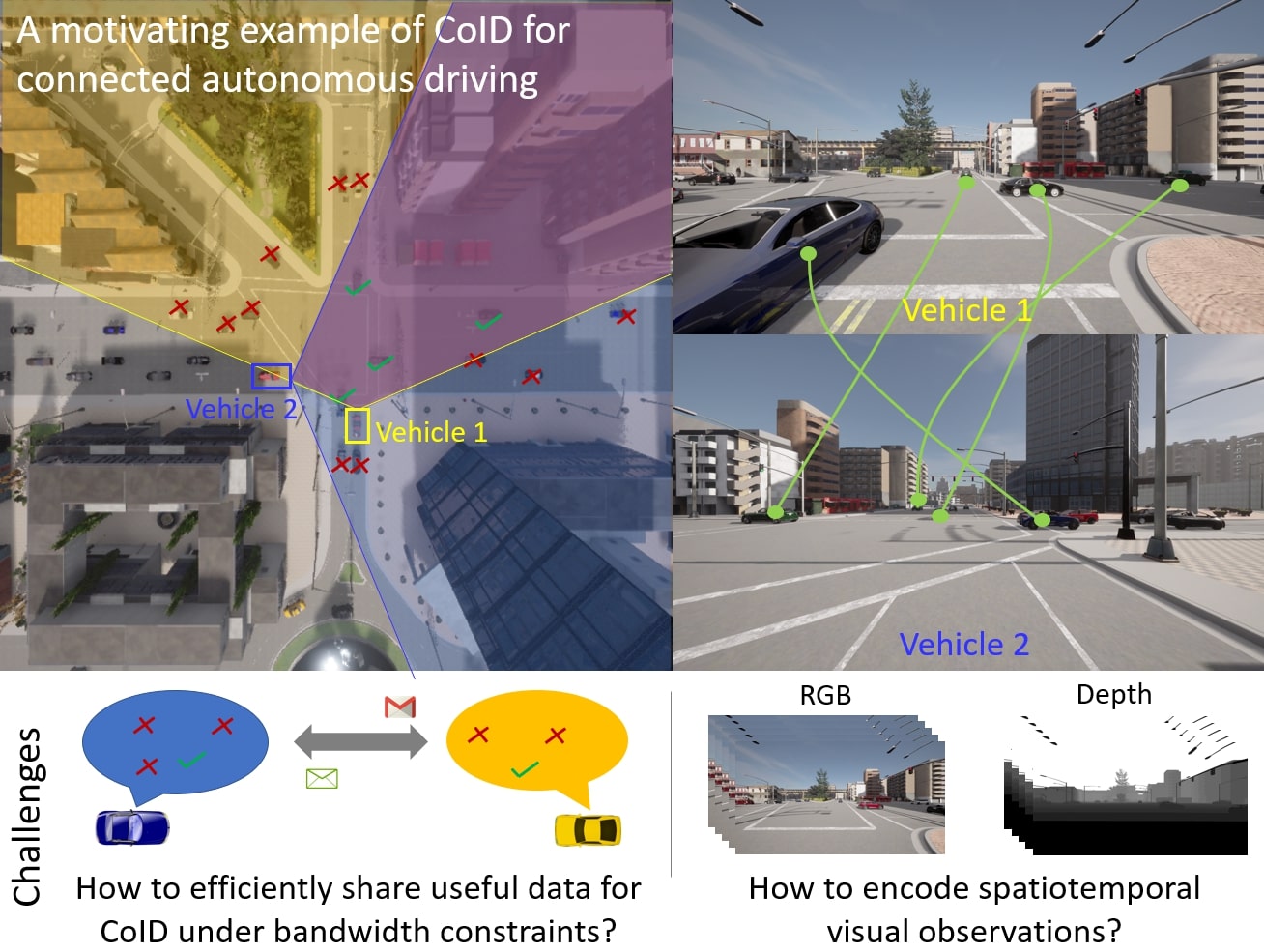}
\caption{A motivating example of CoID under the communication bandwidth constraint for collaborative perception in connected autonomous driving. 
In order to enable connected vehicles to refer to the same street objects,
they must share spatiotemporal information to identify object correspondences, while satisfying the bandwidth constraint.
}
\label{fig:motivation}
\end{figure}

As a critical component of collaborative perception, correspondence identification (CoID) plays a critical role, with the goal of identifying the same objects concurrently observed by multiple robots within their respective field of view. As shown in Figure \ref{fig:motivation}, when a pair of connected vehicles meet at a street intersection, it is important for these vehicles to accurately identify the correspondence of street objects observed in their observations. Given the identified correspondences, connected vehicles can effectively refer to the same objects and estimate their relative poses among robots, thus facilitating further collaboration.

Given the importance of CoID, a variety of techniques are developed,
which fall into two primary categories: learning-free and learning-based approaches. Learning-free techniques comprise keypoint-based visual association \cite{engel2014lsd, he2021transreid}, geometric-based spatial matching \cite{zhang20232, gao2022correspondence}, and synchronization techniques \cite{fathian2020clear, hu2018distributable}.
In contrast, learning-based methods harness deep learning, employing convolutional neural networks (CNNs) for object re-identification from different perspectives \cite{jin2020semantics,khatun2020semantic, voigtlaender2019mots} and graph neural networks (GNNs) for deep graph matching \cite{gao2022asynchronous, fey2019deep}.
Although the previous methods show promising performance, two significant challenges have not been well addressed yet.
The first challenge is caused by the limited communication bandwidth for data sharing in multi-robot systems. In the real-world setting, the maximum bandwidth designated for vehicle-to-everything (V2X) communication is around $7.2$ Mbps \cite{gallo2013short}. It is unrealistic to assume that the available bandwidth allows for the sharing of raw observations, particularly in scenarios involving crowded environments. The second challenge is caused by the need to share temporal data among connected robots,
which further demands a much larger bandwidth compared to sharing single-frame observations.

In order to address these challenges, we propose a novel bandwidth-adaptive method to perform spatiotemporal CoID.
We develop a spatiotemporal graph representation to encode spatiotemporal visual information of street objects observed by each vehicle.
Each node encodes a street object.
Each spatial edge encodes the spatial relationship of a pair of objects,
and each temporal edge is designed to track each object's motion.
Given this graph, 
our approach formulates CoID as a progressive graph-matching problem, while adapting data sharing to the dynamically changing bandwidth.
Specifically, we develop a heterogeneous attention network that generates node features by integrating the objects' visual, spatial, and temporal cues.
We develop a pooling operation that explicitly encodes the spatial and temporal cues to generate a graph-level embedding to encode the global scene.
Then, we introduce our new framework to enforce individual robots to progressively share nodes that are most likely to be also observed by their collaborating vehicles to maximize the CoID performance while satisfying the given communication bandwidth constraints.


The key contribution of this paper is the introduction of the novel bandwidth-adaptive CoID approach for collaborative perception.
Specific novelties include:

\begin{itemize}

\item A novel progressive CoID method for connected vehicles to adapt their data sharing to changing communication bandwidth, which ensures full utilization of the available bandwidth under communication constraints.

    \item A new heterogeneous attention network that integrates visual, spatial, and temporal cues of objects in a unified way,
    while encoding both current and historical cues to enhance the expressiveness of node features for CoID.

    \item A new heterogeneous graph pooling operation to generate a graph-level embedding of the holistic scene, which explicitly encodes the importance of the temporal and spatial cues, as well as compresses the spatiotemporal observations.

\end{itemize}

\section{Related Work}\label{sec:related}
\subsection{Connected Autonomous Driving} The growing interest in connected autonomous driving, driven by collaborative perception among connected agents, has led to various research efforts. Methods are generally classified into raw-based early collaboration, output-based late collaboration, and feature-based intermediate collaboration. Early collaboration fuses raw sensor data from connected agents onboard for vision tasks \cite{arnold2020cooperative}, while late collaboration merges multi-agent perception outputs using techniques like Non-Maximum Suppression \cite{forsyth2014object} and refined matching to ensure pose consistency \cite{song2023cooperative}. Intermediate collaboration strikes a balance by sharing compressed features, with methods such as when2com \cite{liu2020when2com}, who2com \cite{liu2020who2com}, and where2com \cite{hu2022where2comm}. Data fusion strategies include concatenation \cite{chen2019f}, re-weighted summation \cite{guo2021coff}, graph learning \cite{wang2020v2vnet, zhou2022multi}, and attention-based fusion \cite{xu2022v2x, xu2022opv2v}. Applications span object detection \cite{bi2022edge}, tracking \cite{li2021learning}, segmentation \cite{xu2022cobevt}, localization \cite{yuan2022keypoints}, and depth estimation \cite{hu2023collaboration}. However, none of these methods adapts to bandwidth limitations, which often prevent the sharing of holistic information.

\subsection{Correspondence Identification} Correspondence Identification (CoID) methods fall into learning-free and learning-based categories. Learning-free approaches include visual appearance techniques like SIFT \cite{engel2014lsd}, ORB \cite{mur2015orb}, HOG \cite{dalal2005histograms}, and TransReID \cite{he2021transreid}, as well as spatial techniques like ICP \cite{rusinkiewicz2001efficient}, template matching \cite{zhang20232}, and graph matching \cite{gao2020correspondence, gao2021regularized}. Synchronization algorithms also contribute through circle consistency enforcement \cite{fathian2020clear} and convex optimization \cite{hu2018distributable}. Learning-based methods primarily use CNNs \cite{jin2020semantics,khatun2020semantic, voigtlaender2019mots} and GNNs \cite{wang2019learning,zhang2019deep, fey2019deep}, with hybrid approaches like Bayesian CoID \cite{gao2021bayesian, gao2022correspondence} enhancing robustness. However, existing methods struggle to integrate temporal cues, as sharing sequences of frames is constrained by real-world bandwidth limitations. We propose a novel method that integrates visual, spatial, and temporal cues for CoID in a bandwidth-adaptive way.

\begin{figure*}[t]
\vspace{6pt}
\centering
\includegraphics[width=0.9\textwidth]{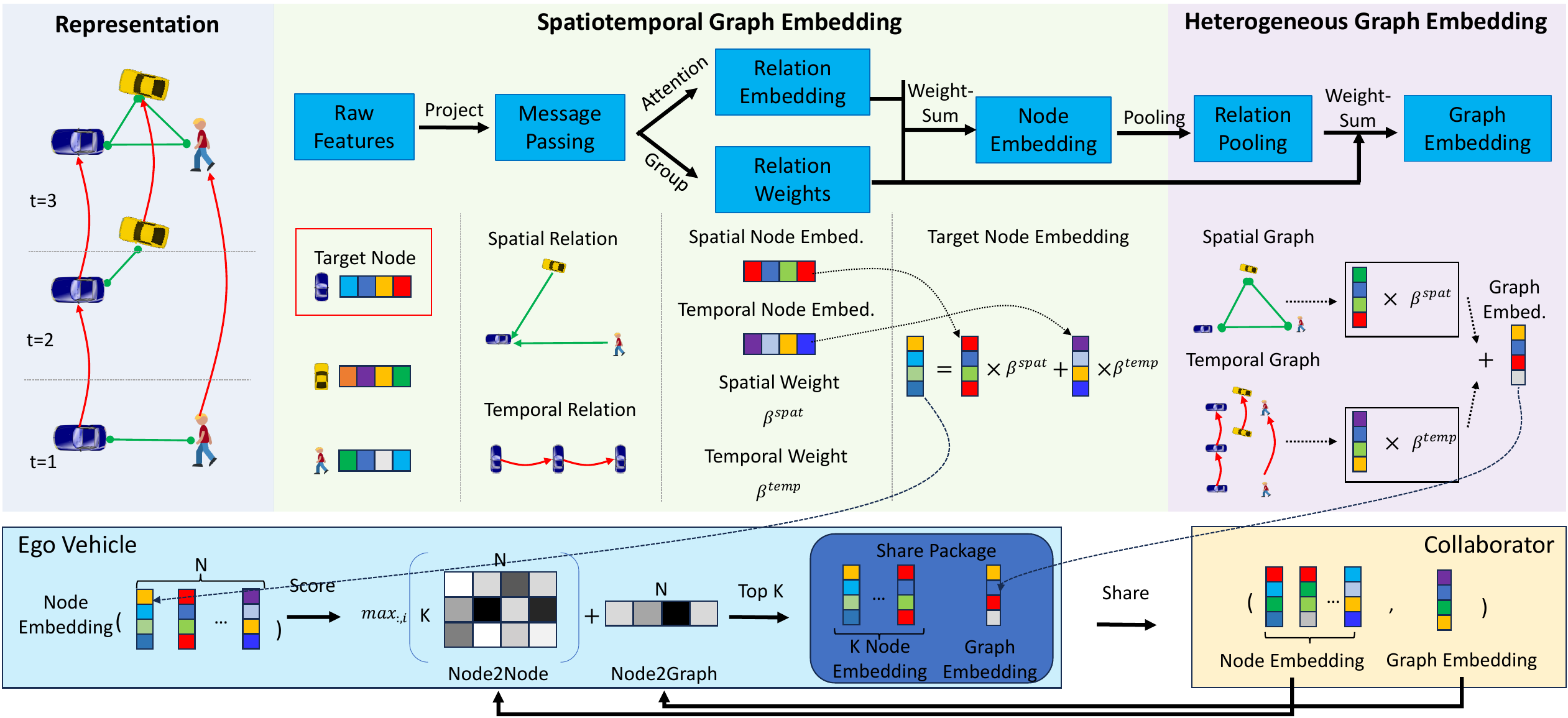}
\caption{
An overview of our proposed bandwidth-adaptive spatiotemporal CoID approach. 
A sequence of observations is represented as a spatiotemporal graph. 
A spatiotemporal graph attention network is used to generate node-level embeddings by integrating spatiotemporal visual cues. 
Then, a heterogeneous graph pooling operation is designed to produce comprehensive graph-level embeddings that explicitly encode the importance of spatial and temporal cues. 
Leveraging both node-level and graph-level embeddings, our approach enables the sharing of node candidates that are likely to be observed by collaborators, 
resulting in effective data sharing that adapts to communication bandwidth constraints.
}
\label{fig:approach}
\end{figure*}
\vspace{-6pt}

\section{Approach}\label{sec:approach}


We discuss our proposed bandwidth-adaptive spatiotemporal CoID method in this section, which is illustrated in Figure \ref{fig:approach}. 
We assume that each of the ego vehicle and collaborative vehicle is equipped with a RGB-D camera or a LiDAR sensor.
Formally, each vehicle obtains a sequence of observations $\OOO = \{obs^t, obs^{t+1}, \dots, obs^{t+T} \}$.
Each observation recorded at time $t$ consists of detected objects $obs^t = \{\vv_1^t, \vv_2^t, \dots, \vv_m^t\}$ where $\vv_i \in \RR^3$ denotes the attributes of the $i$-th object detected at time $t$.
Given a sequence of observations $\mathcal{O}$,
we represent it as a spatiotemporal graph $\GGG=\{\VVV,\EEE^{spat},\EEE^{temp}\}$. $
\VVV$ denotes the node set, which contains all the attributes of objects detected in the observation sequence $\OOO$.
$\EEE^{spat} = \{e_{p,q}^{spat}\}$ denotes the spatial relationships between a pair of
objects.
$e^{spat}_{p,q}$ denotes the distance between the $p$-th object and the $q$-th object recorded at the same time $t$, otherwise $0$.
$\EEE^{temp} = \{e_{p,q}^{temp}\}$ denotes the temporal relationships of the same
object recorded at different times. If $\vv_p^{t_1}$ and $\vv_q^{t_2}$ are the same object recorded at time $t_1$ and $t_2$, then
$e^{temp}_{p,q}=t_2-t_1$, otherwise $0$.


\subsection{Spatiotemporal Graph Embedding}
Given the spatiotemporal graph $\GGG=\{\VVV,\EEE^{spat},\EEE^{temp}\}$, we introduce a new heterogeneous graph attention network that encodes objects' visual, spatial and temporal cues for node-level embedding.
Formally, the node embedding vectors are defined as $\{\mm_i\}^N = \psi(\GGG)$, where  $\psi$ is the heterogeneous attention network. Specifically, we first project each node feature to the same feature space as follows:
\begin{equation}\label{eq:project}
    \hh_i = \WW_v \vv_i
\end{equation}
where $\hh_i$ denotes the projected feature of the $i$-th node, $\WW_v$ denotes the associating weight matrix, and $\vv_i$ denotes the original feature vector of the $i$-th object.
Then we compute the self-attention of each node given different types of edges, defined as follows:
\begin{equation}
    \alpha_{i,j} =  \frac{ \exp\left(\sigma ([\WW\hh_i||\WW\hh_j||\WW_e e_{i,j}])\right)}{\sum_{k\in \NNN^{\Psi}(i)}\exp(\sigma ([\WW\hh_i||\WW \hh_k||\WW_e e_{i,k}]))}
\end{equation}
where $\alpha_{i,j}$ is the attention from node $j$ to node $i$, $\sigma$ denotes the ReLu activation function,  $||$ denotes the concatenation operation, $\WW$ and $\WW_e$ are weight matrices.
This attention $\alpha_{i,j}$ is obtained by comparing the centered $i$-th node with
its neighborhood nodes. Then, we normalize the attention using the SoftMax function.
To encode spatial and temporal relationships of objects, we add edge attributes into the learning process given the edge connection $\NNN^{\Psi}(i)$, where $\Psi \in \{spat, temp\}$ denotes two types of edges connected to the $i$-th node.
Given these two types of edges, we get two attention values $\alpha_{i,j}^{spat}, \alpha_{i,j}^{temp}$.
Then, we compute the node embedding vector as:
\begin{equation}
\hh_i = \sigma \left(\WW\hh_i + \sum_{j \in \NNN^{\Psi}(i)} \alpha_{i,j} (\WW\hh_j +\WW_e e_{i,j}) \right)
\end{equation}
For the same node $i$, we compute its spatial and temporal embedding vectors $\hh_i^{spat}$ and $\hh_i^{temp}$ given its associated attentions $\alpha_{i,j}^{spat}, \alpha_{i,j}^{temp}$. They are computed via aggregating the node and edge embedding features weighted by attention values.
We also use a multi-head mechanism to enable the network to catch a richer representation of the embedding. Multi-head embedding vectors are concatenated after intermediate attention layers.

To combine these two spatial and temporal embedding vectors of the same node, we learn the weights of spatial and temporal relationships of nodes to indicate their importance,
which is defined as follows:
\begin{equation}\label{eq:beta_spat}
    \beta^{spat} = \frac{1}{|\VVV^{spat}|}\sum_{i \in \VVV^{spat}} \qq^T \tanh(\WW_b\hh_i +\bb)
\end{equation}
\begin{equation}\label{eq:beta_temp}
    \beta^{temp} = \frac{1}{|\VVV^{temp}|}\sum_{i \in \VVV^{temp}} \qq^T \tanh(\WW_b\hh_i +\bb)
\end{equation}
where $\WW_b$ denotes the weight matrix, $\bb$ is the bias vector, $\qq$ denotes the learnable edge-specific attention vector. The learnable parameters are shared for all spatial and temporal relationships. Then, the spatial and temporal weights  are normalized through SoftMax, defined as follows:
\begin{equation}
    \beta^{spat} = \frac{\exp(\beta^{spat})}{\exp(\beta^{spat} + \beta^{temp})}
\end{equation}
\begin{equation}
    \beta^{temp} = \frac{\exp(\beta^{temp})}{\exp(\beta^{spat} + \beta^{temp})}
\end{equation}
where $\beta^{spat}, \beta^{temp}$ denote the weights of spatial and temporal relationships for CoID. The higher the value, the larger the importance of the type of relationship. The final node embedding is computed as:
\begin{equation}\label{eq:final}
\mm_i = \sum_{\Psi \in \{spat, temp\}} \beta^\Psi \hh^\Psi_i
\end{equation}
where $\mm_i$ denotes the final embedding vector of the $i$-th object in the spatiotemporal graph by integrating object positions, and spatiotemporal relationships.

\subsection{Heterogeneous Graph Pooling}
Due to the communication bandwidth constraint, the collaborator robot can only share partial nodes with the ego robot. To also share the comprehensive information of the collaborator's observations for CoID, we further propose a novel graph pooling operation, which compresses the collaborator's spatiotemporal graph into a single vector and integrates with the proposed heterogeneous graph network in a principled way.

Specifically, we decompose the graph $\GGG=\{\VVV,\EEE^{spat},\EEE^{temp}\}$ into two separate graphs, including  $\GGG^{spat}=\{\VVV,\EEE^{spat}\}$ and $\GGG^{temp}=\{\VVV,\EEE^{temp}\}$. 
Then, we perform ASAPooling \cite{ranjan2020asap} on both of the spatial and temporal graphs, which is defined as follows:
\begin{equation}\label{eq:pooling}
    \zz^\Psi = \phi(\{\mm_i\}^N, \EEE^\Psi), \Psi = \{spat, temp\}
\end{equation}
where $\phi$ is the ASAP pooling operation, and $\zz^\Psi = \{\zz^{spat}, \zz^{temp}\}$ is the graph-level embedding concerning the spatial and temporal edges. By taking advantage of the weights of spatiotemporal relationships, we compute the final graph embedding vector as:
\begin{equation}\label{eq:graphembed}
    \zz = \beta^{spat}\zz^{spat} + \beta^{temp}\zz^{temp}
\end{equation}
where $\zz$ denotes the graph embedding vector of the spatiotemporal graph $\GGG$, which is computed by the sum of spatial and temporal embedding vectors weighted by the importance of different types of relationships.

\subsection{Bandwidth-Adaptive CoID}
Given the node and graph-level embedding vectors generated by each robot, we design a novel interactive approach to perform CoID. 
Specifically, on the ego robot side, we have ego node embedding vectors $\{\mm\}^N$, collaborator node embedding vectors $\{\mm^\prime\}^{K}$ and the graph embedding vector $\zz^\prime$. $K$ denotes the communication bandwidth that allows the maximum number of nodes to share. Then we compute matching scores to indicate the probabilities of ego nodes to appear in the collaborator's observations. The computation is defined as:
\begin{align}\label{eq:sim}
    \SIM^{node}_{i,j} &= \exp(-||\mm_i-\mm_j^\prime||_2) \nonumber\\
    \sss^{graph}_{i} &= \exp(-||\mm_i-\zz^\prime||_2)
\end{align}
where $\exp()$ denotes the exponential operator, 
$\SIM^{node} \in \mathbb{R}^{N \times K}$ represents the similarity between pairs of ego-collaborator node embedding vectors and
$\sss^{graph} \in \mathbb{R}^{N}$ denotes the similarity between the ego node embedding vector and collaborator graph embedding vector. Finally, we select the $Top_{K}$ candidates given $\{\SIM^{node}_{i,j}\}$ and $\{\sss^{graph}_{i}\}$. Specifically, 
\begin{equation}\label{eq:matching_score}
    \sss = \lambda \sss^{graph} + (1-\lambda) \max_{:, i} \SIM^{node}, \quad i = 1,2,\dots, N
\end{equation}
where $\sss$ denotes the final matching score to select correspondence candidates, $\lambda$ denotes a hyperparameter to indicate the importance of node2node similarity and node2graph similarity, and $\max_{:, i} \SIM^{node}$ denotes the maximum elements in each column of $\SIM^{node}$. Then, the top-K candidates are selected as $\{\mm_i\}^K = Top_K (\sss)$ and share with the collaborator. This process interactively runs on both ego and collaborator vehicles until reaching the maximum interaction number. 

Given the received candidate set $\MMM = \{\mm^\prime\}^m$, where $m$ is the total number of nodes sent from the collaborator and the ego robot's graph $\GGG$, we perform graph matching to identify the correspondences of objects detected in multi-robot observations. Specifically, we compute the similarity  $\AAAA \in \mathbb{R}^{n \times m}$ by:
\begin{equation}\label{eq:similarity}
    \AAAA_{i,j} =\mm_i^{ \top}\mm_j^{\prime} 
\end{equation}
where the similarity score $\AAAA$ contains
similarities of spatiotemporal visual features of the objects,
$\mm_i \in \GGG$ denotes the node embedding vectors in the ego graph $\GGG$,
and $\mm_j^\prime \in \MMM$ denotes the node embedding vectors shared by the collaborator.  
The correspondences can be identified as 
 $\YY=\text{SoftMax}(\AAAA)$,
where  $\YY$ is the correspondence matrix with $\YY_{i,j}=1$ denoting the correspondence between the $i$-th object in $\GGG$ and the $j$-th object in $\MMM$, otherwise $\YY_{i,j}=0$.
We use the circle loss to train our network, which is defined as:
\begin{align}\label{eq:loss}
     L_{\GGG' \rightarrow \GGG} (\DD) &= \sum_{\vv^\prime_i \in \VVV^\prime} \log \left(1+\sum_{\vv_j \in \VVV^{p}} \exp \left[\gamma(\DD_{i,j} - \delta_p)^2\right] \right.  \nonumber \\
     & \left. + \sum_{\vv_k \in \VVV^{n}} \exp\left[\gamma(\delta_n - \DD_{i,k})^2\right]\right)
\end{align}
\normalsize
$L_{\GGG' \rightarrow \GGG}$ describes the loss given node sets  $\VVV = \{\VVV^p, \VVV^n\}$ and $\VVV'$. $\VVV^p$ denotes the positive nodes that have corresponding nodes in graph $\GGG'$. $\VVV^n$ denotes the negative nodes that have no corresponding nodes in graph $\GGG'$.
$\delta_p$ and $\delta_n $ are two hyperparameters, which denote the positive and negative margins separately. $\gamma$ denotes the scale factor.
$\DD = \{D_{i,j}\}^{n \times n'}$ denotes the distance matrix. We design two different distance matrices in this paper, including $\DD_{i,j}^{node} = ||\mm_i -\mm'_j||_2$ denoting the distance between a pair of node feature vectors, and $\DD_{i}^{graph} = ||\mm_i -\zz||_2$ denoting the distance between the node feature and graph feature. 
Similar to the definition of $L_{\GGG' \rightarrow \GGG}$, we can compute the loss $L_{\GGG \rightarrow \GGG'}$ given node sets $\VVV$ and $\VVV' = \{\VVV^{\prime p}, \VVV^{\prime n}\}$.
The final loss is:
\begin{align}
    L &= \frac{1}{4} \left(L_{\GGG \rightarrow \GGG'} (\DD^{node}) + L_{\GGG' \rightarrow \GGG}(\DD^{node}) \right.\nonumber \\
    & \left. + L_{\GGG \rightarrow \GGG'} (\DD^{graph}) + L_{\GGG' \rightarrow \GGG}  (\DD^{graph} ) \right)
\end{align}
\normalsize
The full algorithm is presented in the supplementary \footnote{Supplementary: \url{https://gaopeng5.github.io/acoid}}.

\section{Experiment}\label{sec:experiment}

\subsection{Experimental Setups}
We have developed a high-fidelity connected autonomous driving (CAD) simulator by seamlessly integrating two open-source platforms: CARLA \cite{Dosovitskiy17} and SUMO \cite{krajzewicz2002sumo}. In the simulations, we collect data to train and evaluate our proposed approach. The details are in the supplementary  \footnotemark[\value{footnote}].

\begin{figure*}[htb]
	\centering
\subfigure[DGMC]{\includegraphics[height=4.0cm]{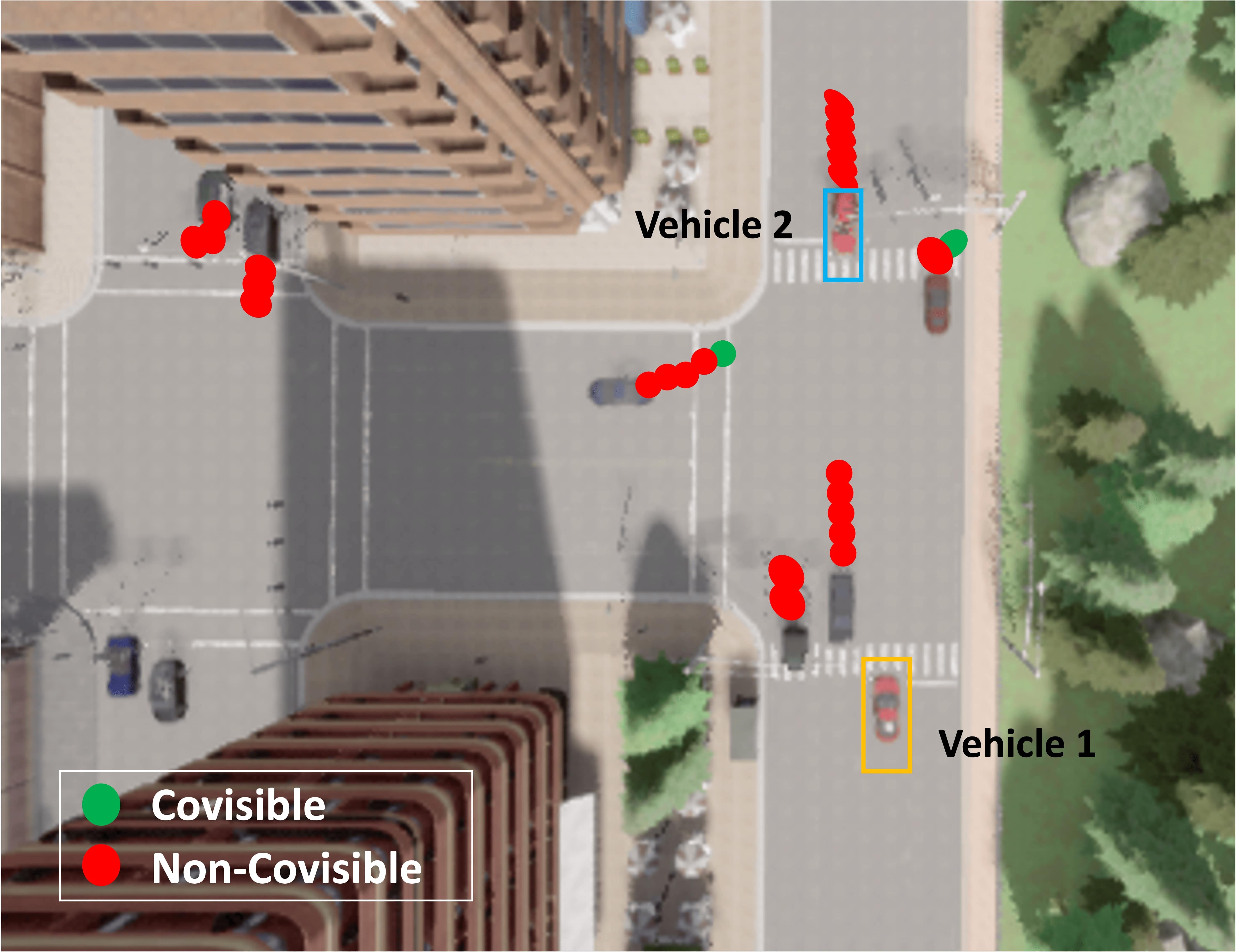}}
\subfigure[Ours]{\includegraphics[height=4.0cm]{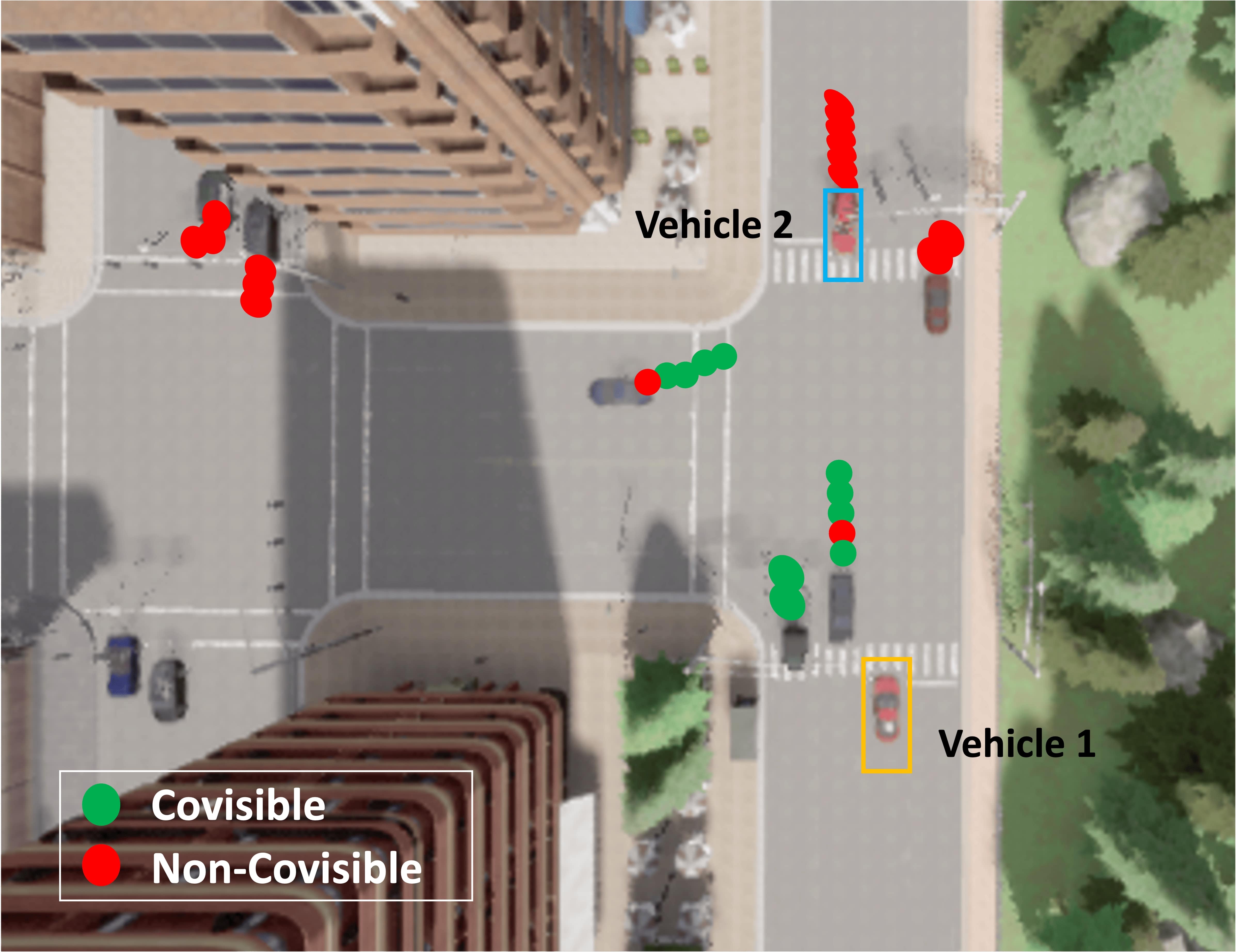}}
\subfigure[GroundTruth]{\includegraphics[height=4.0cm]{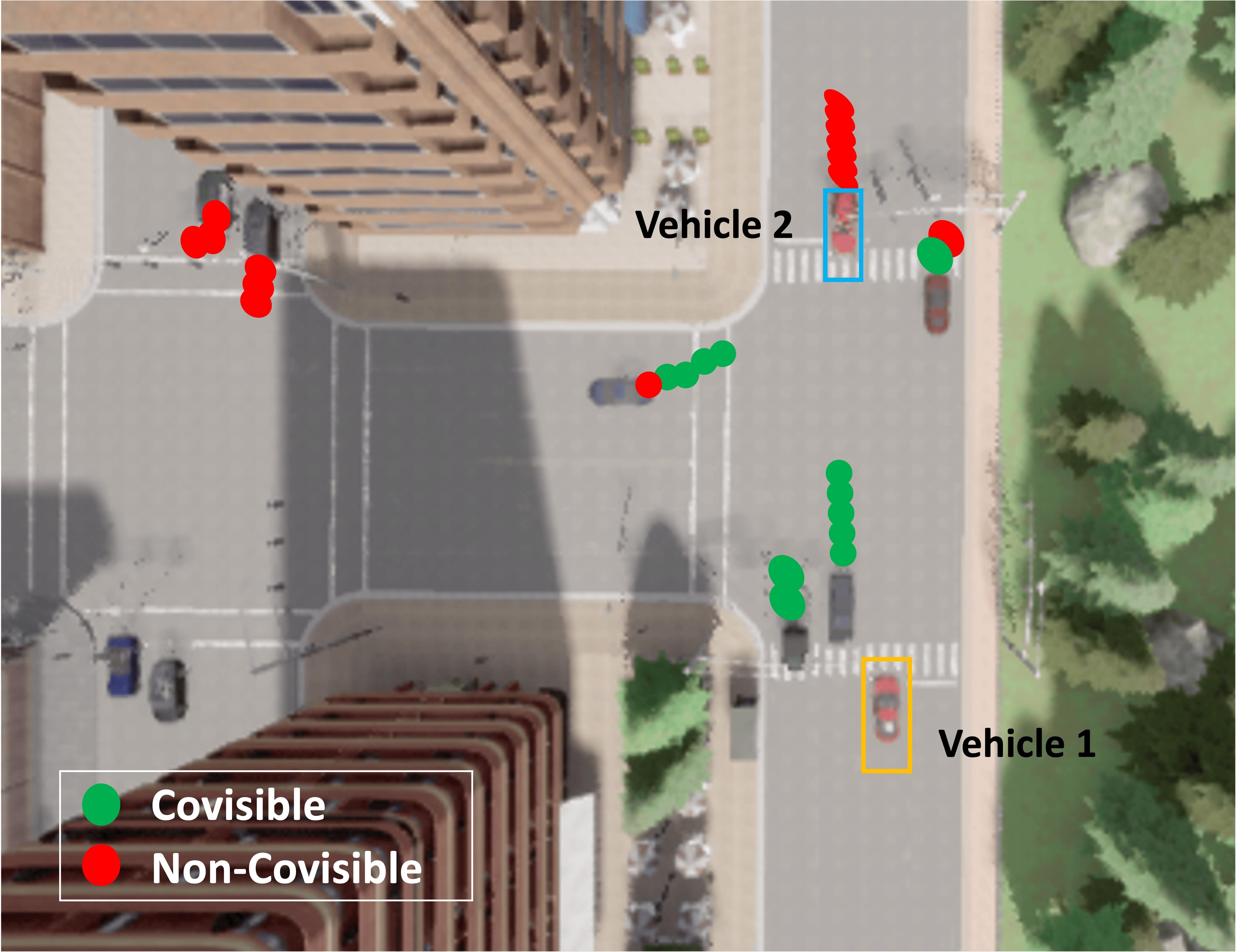}}
\subfigure[DGMC]{\includegraphics[height=4.0cm]{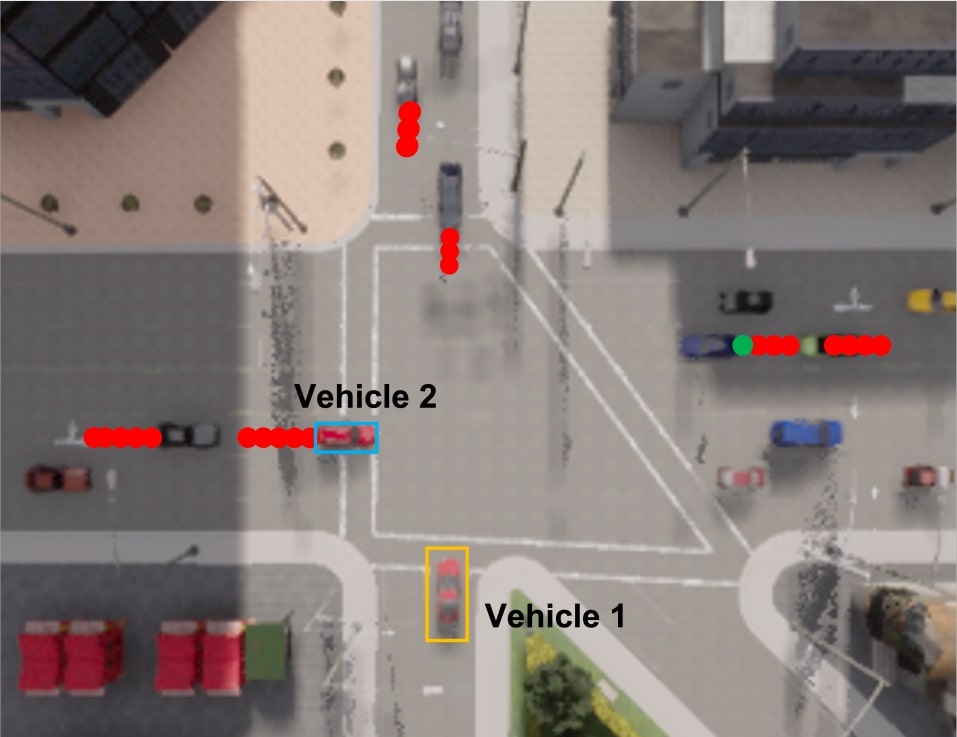}}
\subfigure[Ours]{\includegraphics[height=4.0cm]{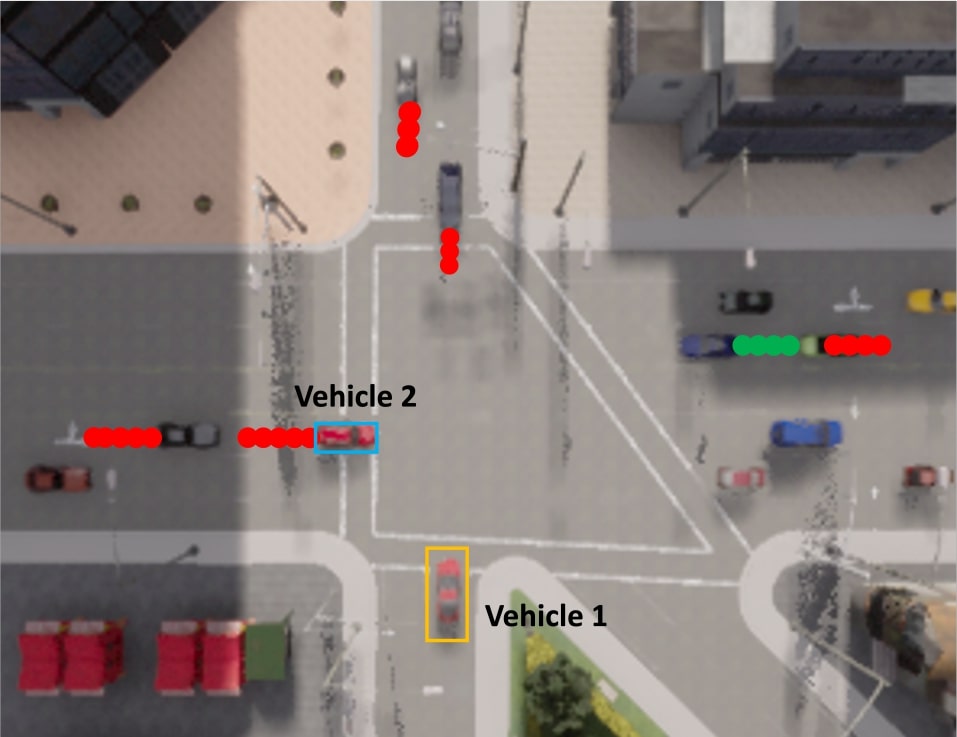}}
\subfigure[GroundTruth]{\includegraphics[height=4.0cm]{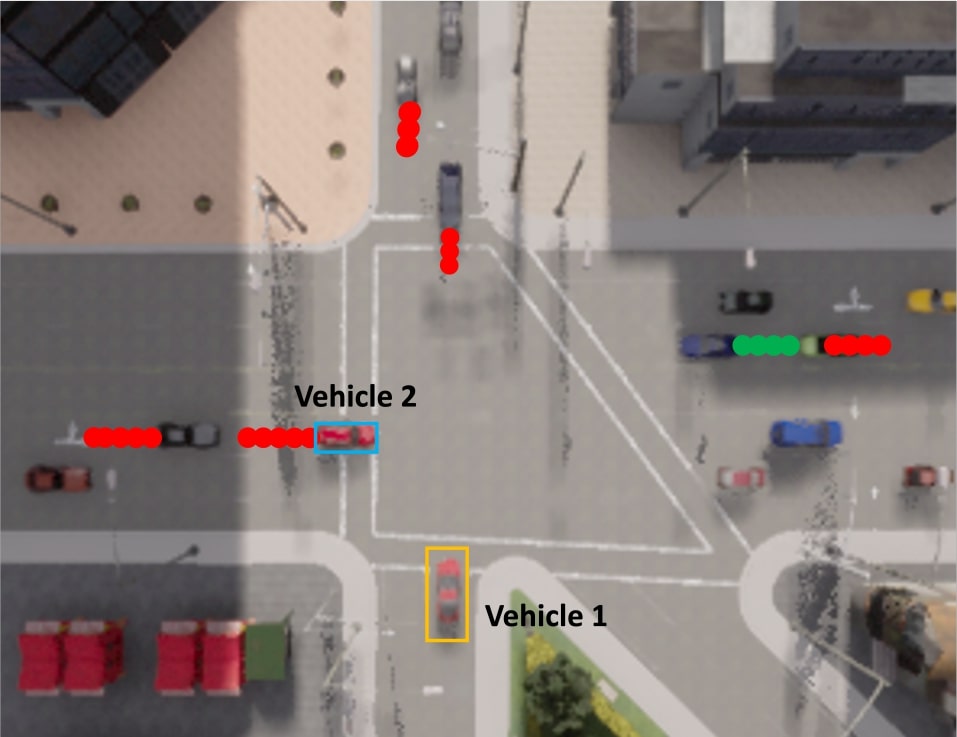}}
\vspace{-6pt}
\caption{Qualitative  results obtained by our approach in normal traffic scenarios (the first row) and crowd traffic scenarios (the second row),
as well as comparisons with the DGMC method and the ground truth. 
The sequence of points denotes a history of observations, which consists of 1-5 points indicating object locations in the past 1-5 time steps.
Red points denote non-covisible objects that can only be observed by Vehicle 1. 
Green points represent the identified covisible objects that can be observed by both Vehicle 1 (in the orange bounding box) and Vehicle 2 (in the blue bounding box). 
All results are computed in the setup that Vehicle 1 receives information from Vehicle 2, and Vehicle 2 aims to share covisible objects with Vehicle 1. }\label{fig:qual}
\end{figure*}

\begin{table*}[ht]
\centering
\tabcolsep=0.15cm
\caption{Quantitative results in the CAD simulations. The BIS metric is used to evaluate recall and communication efficiency of CoID. }
\label{tab:QuanResults}
\begin{tabular}{|c|c|c|c|c|c|c|c|c|c|c|c|c|c|c|c|c|c|}
\hline
Scenario &\multicolumn{3}{ c|}{ 1 (Normal)}  &\multicolumn{3}{ c|}{ 2 (Normal)  } &\multicolumn{3}{ c|}{ 3 (Normal)}  &\multicolumn{3}{ c|}{ 4 (Crowd)}
\\
\hline
 Method
& Precision $\uparrow$& Recall $\uparrow$& BIS $\uparrow$
& Precision $\uparrow$& Recall $\uparrow$& BIS $\uparrow$
& Precision $\uparrow$& Recall $\uparrow$& BIS $\uparrow$
& Precision $\uparrow$& Recall $\uparrow$& BIS $\uparrow$  \\
\hline

\hline
{GCN-GM} \cite{fey2018splinecnn}
& 0.5457 & 0.5294 & 1.1056
& 0.3771& 0.7074 &1.6801
&0.5441 & 0.5373 & 1.4359
& 0.3923 &0.5724 & 1.9737\\

BDGM \cite{gao2021bayesian}
& 0.5420 & 0.5345 & 1.1222
& 0.3538 & 0.6806 & 1.6163
& 0.5488 & 0.5296 & 1.4153
& 0.3852 & 0.5624 & 1.9390\\

DGMC \cite{fey2019deep}
& 0.5588 & 0.5518 & 1.1331
& 0.3665 & 0.7052 & 1.6747
& 0.5544 & 0.5397 & 1.4421
& 0.3896 & 0.5716 & 1.9709\\

Ours-NE
&0.5995 & 0.6968 & 1.3993
& 0.4867 & 0.8851 & 2.4345
& 0.6478 & 0.8890 & 1.5290
& 0.4503 & 0.6166& 2.2979\\
Ours
&\textbf{0.6102} & \textbf{0.7123} &\textbf{1.4305}
&\textbf{0.5044} & \textbf{0.9241} & \textbf{2.5418}
&\textbf{0.6765} & \textbf{0.9261} & \textbf{1.5291}
& \textbf{0.4737} & \textbf{0.6756} & \textbf{2.5008}\\
\hline
\end{tabular}
\end{table*}

We implement the full version of our method for bandwidth-adaptive spatiotemporal CoID,
which uses $Top_{K}(\sss)$ to select candidates for CoID.
In addition, we implement a baseline method labeled as \textbf{Ours-NE}, 
which uses $Top_{K}(\SIM^{node})$ to only compare node embeddings (NE) of the collaborators and the ego vehicle to select candidates.
Furthermore, we compare our approach with three previous CoID methods, including:
\begin{itemize}
    \item Graph convolutional neural network for graph matching (\textbf{GCN-GM}) that uses the spline kernel to aggregate visual-spatial information of  objects for CoID \cite{fey2018splinecnn}.
    \item Deep graph matching consensus (\textbf{DGMC}) that performs an iterative refinement process on the similarity matrix given the consensus principle \cite{Fey2020}.
    \item Bayesian deep graph matching (\textbf{BDGM}) that performs CoID under a Bayesian framework to reduce non-covisible objects with high visual uncertainty \cite{gao2021bayesian}.
\end{itemize}
These comparison methods use randomly selected query nodes sent from collaborators.
None of the comparison methods are capable of integrating temporal information and addressing communication bandwidth adaptation issues.

To quantitatively evaluate the CoID performance, we employ the following metrics:
\begin{itemize}
    \item \textbf{Precision} is defined as the ratio of the retrieved objects with correspondences over all retrieved correspondences.
    \item \textbf{Recall} is defined as the ratio of the retrieved objects with correspondences over the ground truth correspondences.
    \item \textbf{F1 Score} is to evaluate the overall performance of CoID methods, which is defined as:$\frac{2\times Precision \times Recall}{Precision +Recall}$
    \item \textbf{BIS}, following the recent work \cite{liu2020who2com}, it is defined as: $Recall * \frac{\# Query}{\# Interact * \# Shared}$,
    where $\# Query$ is the number of query nodes (collaborator's observed nodes), $\# Interact$ denotes the times of interaction, and $\# Shared$ denotes the number of shared nodes in each interaction.
    The second term describes the ratio of shared nodes compared with all query nodes. Thus, a higher {BIS} value indicates a higher recall and a lower amount of shared data.
    {BIS}  $=1$ means that the collaborator shares all of its observations with the ego vehicle and the recall is $1$.
\end{itemize}

\subsection{Results in Normal Traffic Scenarios}
We designed three normal traffic scenarios (Scenarios 1, 2, and 3) in the CAD simulation to evaluate our approach. These scenarios cover key challenges like complex street object interactions, occlusion, limited communication bandwidth, missing objects in observation sequences, and dynamic traffic with pedestrians and vehicles. Our approach runs on a Linux machine with an i7 16-core CPU, 16GB RAM, and an RTX 3080 GPU. It processes at around 60 Hz, with spatiotemporal graph generation at 300 Hz.

The qualitative results in normal traffic scenarios are shown in Figure \ref{fig:qual}. In these results, vehicle 1 receives query nodes from vehicle 2, and after CoID computation, vehicle 1 selects covisible objects (in green) to share back with vehicle 2. Non-covisible objects (in red) are outliers and should not be shared, as they negatively impact CoID performance and increase bandwidth usage. Our interactive CoID method effectively retrieves covisible objects by preserving the spatiotemporal relationships of street objects in a communication-efficient manner.
The quantitative results in Table \ref{tab:QuanResults} show that our approach outperforms all the other methods in three scenarios in normal traffic. It is due to our approach’s ability to integrate multi-vehicle spatial and temporal observations. Our approach achieves a significant improvement of $8\%-56\%$ in covisible object retrieval and data-sharing efficiency based on the BIS metric, making it ideal for real-world applications with communication constraints. Additionally, our baseline method performs much better than other comparisons, highlighting the importance of temporal cue integration. The full approach surpasses the baseline due to our novel heterogeneous graph pooling and interactive CoID mechanism.

\begin{figure}[t]
\centering
\subfigure[DGMC]{\includegraphics[height=3.0cm]{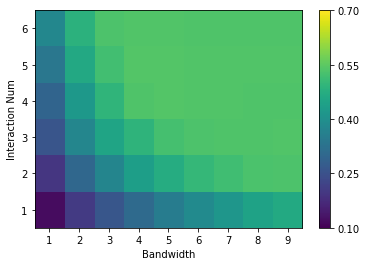}\label{fig:dgmc}}
\subfigure[Ours]{\includegraphics[height=3.0cm]{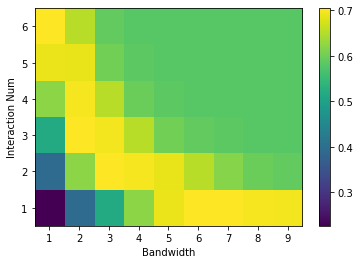}\label{fig:ours}}\vspace{-6pt}
\caption{Comparison between DGMC and our approach on the F1 score given different number of iterations and bandwidth constraints.}\label{fig:compare}
\vspace{-8pt}
\end{figure}

In Scenario 2, we conducted experiments to deeply compare our method with the existing DGMC method based on the F1 score, as depicted in Figure \ref{fig:compare}. The x and y axes represent the bandwidth and interaction number between connected vehicles.
Our approach demonstrates superior performance compared to DGMC in both effectiveness and efficiency. From an effectiveness standpoint,  both methods show gradual improvement with increasing bandwidth and interaction numbers. This is because of the increase of received query nodes from collaborators, which leads to the increase of recall. However,
our approach consistently achieves a significantly higher F1 score compared to DGMC with the same number of interactions and bandwidth.
From an efficiency standpoint, our approach attains its best performance by sharing/receiving around 10 nodes, as indicated by the yellow regions. In contrast, the DGMC method requires over 30 nodes to achieve a performance inferior to ours. As the vehicles start to share outliers in this case, it leads to a decrease in accuracy.

\subsection{Results in Crowded Traffic Scenarios}
We introduce a challenging Scenario 4 to thoroughly assess our approach. In contrast to Scenarios 1, 2, and 3, Scenario 4 features a higher density of entities, hosting over 20 street objects within a street intersection. Employing a sequence of frames to generate the spatiotemporal graph representation results in a graph with over 50 nodes. Scenario 4 introduces extensive interactions among street objects, significantly amplifying the complexity of the CoID task.

The quantitative results for crowded traffic scenarios in Table \ref{tab:QuanResults} show that our approach outperforms all metrics, thanks to the heterogeneous graph pooling technique, which efficiently selects covisible candidates from dense graphs. We achieve over a 25$\%$ improvement in covisible object retrieval and data-sharing efficiency (BIS metric), demonstrating the robustness of our approach in both normal and crowded scenes. Our baseline, Ours-NE, also surpasses other methods by integrating spatiotemporal cues for CoID. For the qualitative results, Figure \ref{fig:qual} highlights objects observed by vehicle 1, showing that our method retrieves more covisible objects than DGMC, emphasizing the importance of spatiotemporal cue integration and interactive node sharing.


\subsection{Discussion}
\begin{figure}[t]
	\centering
\subfigure[Temporal cues]{\includegraphics[height=3.1cm]{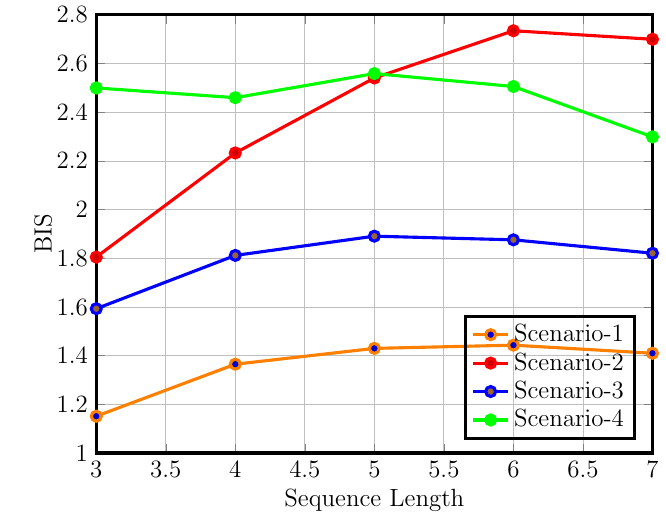}\label{fig:temp}}
\subfigure[CoID]{\includegraphics[height=3.1cm]{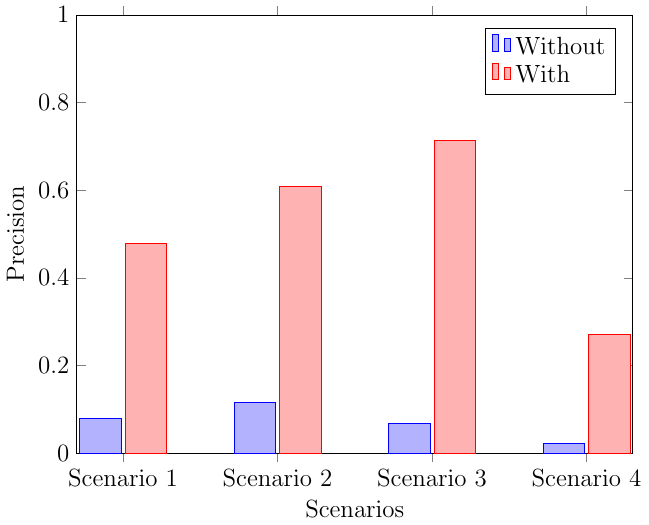}\label{fig:coid}}
\vspace{-6pt}
\caption{Analysis of our approach's characteristics in the CAD simulation, including the effect of the length of temporal sequence based on BIS and the improvements of CoID based on precision.
}\label{fig:analysis}
\vspace{-8pt}
\end{figure}

Figure \ref{fig:temp} depicts that the performance of our approach gradually decreases as the sequence length increases. When the sequence length is in the range of $[5,6]$, the best performance is achieved. If the sequence length keeps increasing, the performance becomes stable with a small fluctuation, which also indicates the effectiveness of our approach which can capture the temporal cues for CoID.


Figure \ref{fig:coid} presents the improvements of CoID accuracy by using our approach to retrieve covisible objects. We use our approach to retrieve covisible objects and then use the traditional graph-matching approach to identify the correspondences of objects. Without using our approach as a pre-process, the graph-matching performance is bad due to the existence of a large amount of outliers. Given the retrieved objects obtained by our approach, the CoID accuracy improves significantly, as our approach first shares the objects that are most likely to have correspondences, thus reducing a large number of outliers.


\section{Conclusion}\label{sec:conclusion}
In this paper, we introduce a novel bandwidth-adaptive spatiotemporal CoID approach for collaborative perception. We formulate CoID as a spatiotemporal graph learning and matching problem, developing an interactive information-sharing paradigm that adapts to dynamic bandwidth constraints. Our method uses a heterogeneous graph attention network to integrate visual, spatial, and temporal features and employs graph pooling to select candidates for data sharing. Extensive experiments in both normal and crowded traffic simulations demonstrate that our approach achieves state-of-the-art CoID performance under varying bandwidth conditions.

\bibliographystyle{IEEETran}
\bibliography{main}

\end{document}